\documentclass[letterpaper, 10 pt, conference]{ieeeconf}  

\IEEEoverridecommandlockouts                              

\usepackage{cite}
\usepackage{amsmath,amssymb,amsfonts}
\usepackage{algorithmic}
\usepackage{graphicx}
\usepackage{textcomp}
\usepackage{xcolor}

\usepackage{subcaption}

\DeclareMathOperator*{\argmin}{arg\,min}

\title{\LARGE \bf
An Adversarial Objective for Scalable Exploration
}

\author{
Bernadette Bucher$^{1}$*, Karl Schmeckpeper$^{1}$*, Nikolai Matni$^{1}$, Kostas Daniilidis$^{1}$
\thanks{* Alphabetical ordering; the first two authors contributed equally.}
\thanks{$^{1}$GRASP Laboratory, University of Pennsylvania
{\tt\small \{bucherb, karls, nmatni, kostas\}@seas.upenn.edu}
}
}

\begin{document}

\maketitle
\thispagestyle{empty}
\pagestyle{empty}

\begin{abstract}
Model-based curiosity combines active learning approaches to optimal sampling with the information gain based incentives for exploration presented in the curiosity literature. Existing model-based curiosity methods look to approximate prediction uncertainty with approaches which struggle to scale to many prediction-planning pipelines used in robotics tasks. We address these scalability issues with an adversarial curiosity method minimizing a score given by a discriminator network. This discriminator is optimized jointly with a prediction model and enables our active learning approach to sample sequences of observations and actions which result in predictions considered the least realistic by the discriminator. We demonstrate progressively increasing advantages as compute is restricted of our adversarial curiosity approach over leading model-based exploration strategies in simulated environments. We further demonstrate the ability of our adversarial curiosity method to scale to a robotic manipulation prediction-planning pipeline where we improve sample efficiency and prediction performance for a domain transfer problem.
\end{abstract}


\section{Introduction}

\begin{figure}
    \centering
    \vspace{0.2cm}
    \includegraphics[width=\columnwidth]{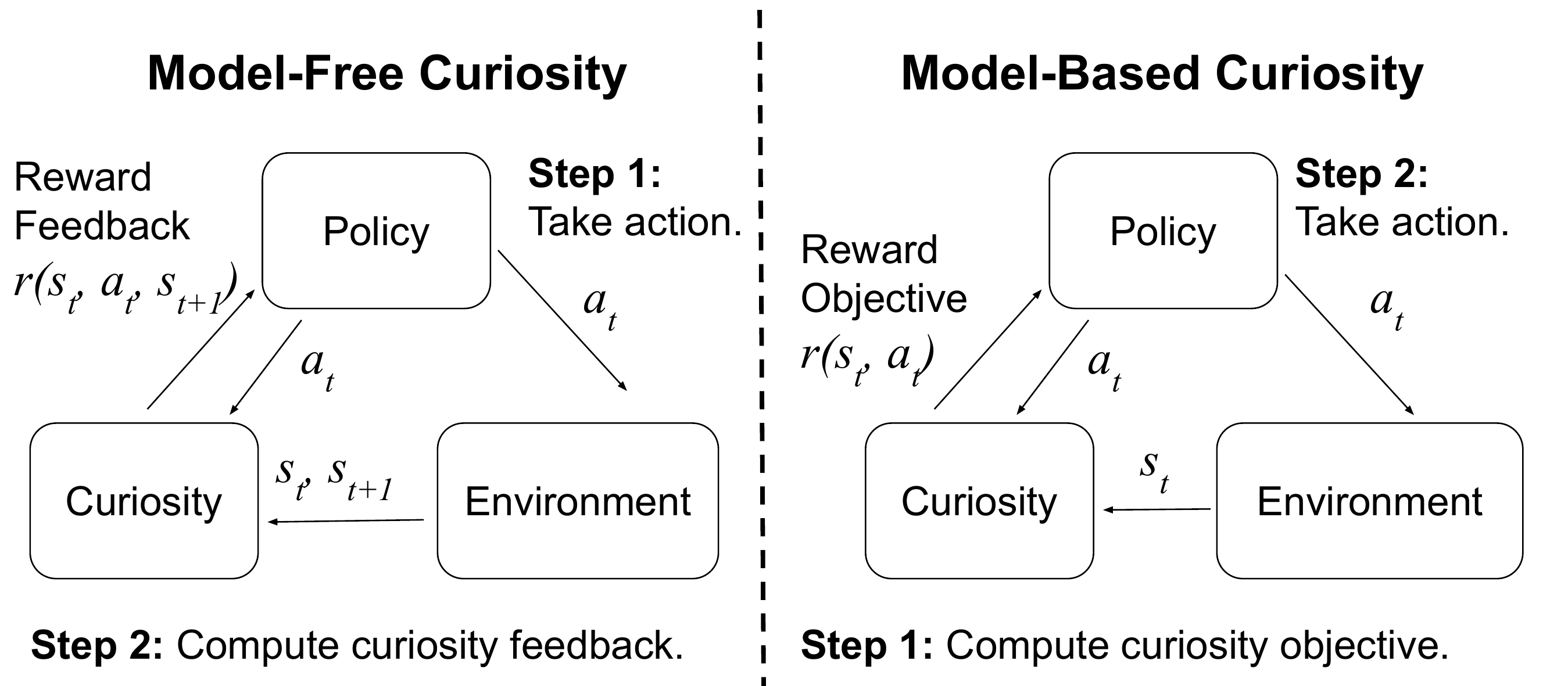}
    \caption{Our model-based curiosity approach contrasted to established model-free formulations of curiosity.  Existing model-free approaches to curiosity calculate the curiosity score after the action has been taken, while our approach uses the curiosity score to decide which action to take.}
    \label{fig:curiosity}
    \label{fig:free_the_models}
    \vspace{-0.2cm}
\end{figure}

Methods for \textit{curiosity} maximize expected information gain of predictive models (typically via informal mathematical proxies) which can be used to perform targeted sampling \cite{funTheory, boredom, jaegle2019visual}. Model-free curiosity derives rewards \textit{after} an action is taken and thus requires knowledge of the action outcome. This approach necessitates integration with model-free reinforcement learning in which rewards provide feedback to an updated policy for selecting actions. In contrast, model-based methods are active learning strategies which use a prediction model directly to select actions, so curiosity measurements must be made \textit{before} the action is taken to execute curious behavior. The fundamental difference between these approaches is visualized in Figure \ref{fig:free_the_models}.

\begin{figure}
    \centering
    \includegraphics[width=0.8\columnwidth]{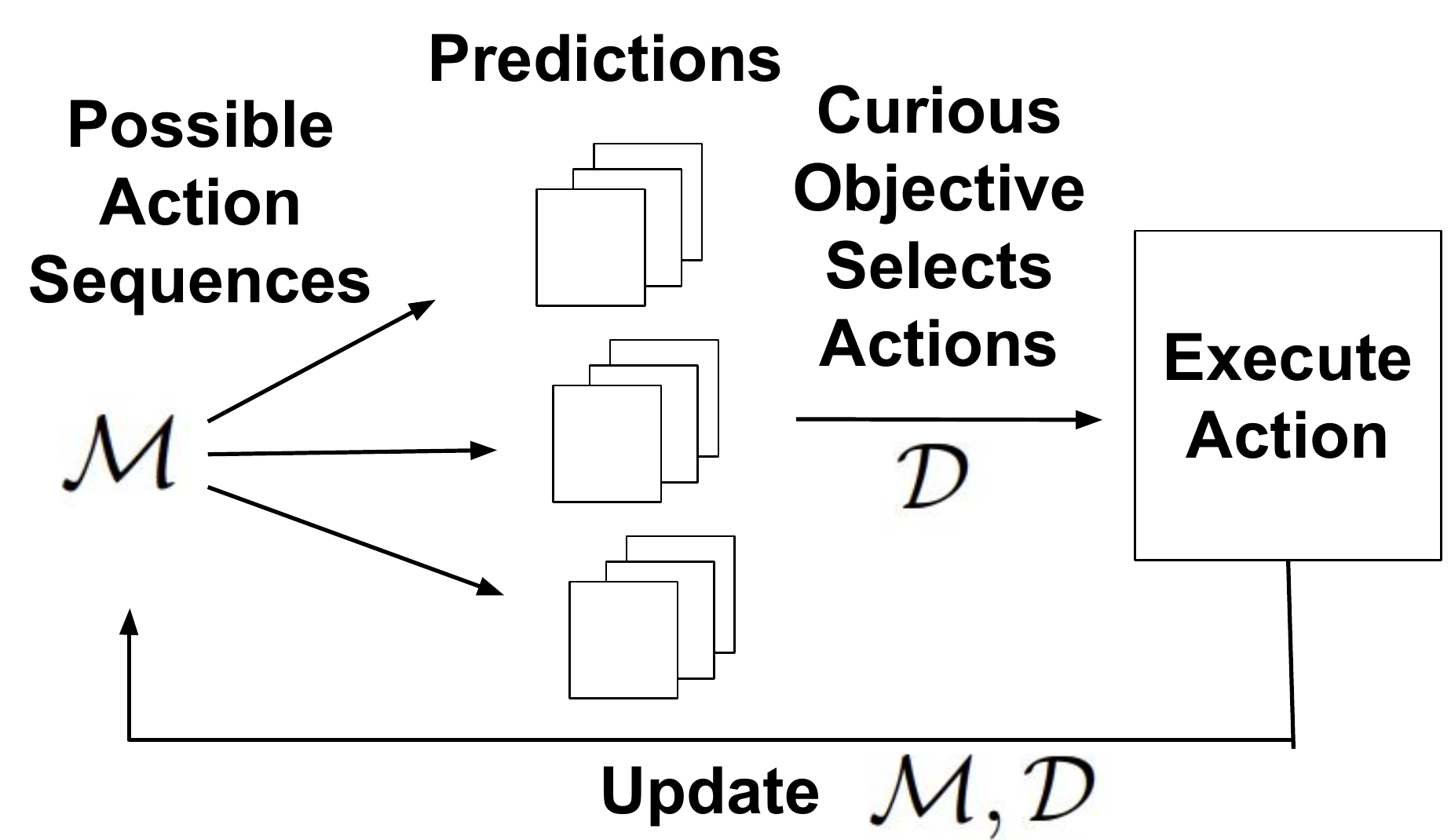}
    \caption{Our approach for model based curiosity.  A predictive model generates predictions on a number of potential trajectories.  These predictions are evaluated with a discriminator and trajectory that corresponds to the least realistic prediction is executed.  The predictive model and the discriminator are updated with the newly collected data.}
    \label{fig:online_method_diagram}
\end{figure}

The active learning and active perception literature has long established the ability of good sampling strategies to increase sample efficiency and model performance in robotics \cite{roy2001toward, Cohn1996, Bajcsy1992}. Model-based curiosity research distinguishes itself within this broader set of work by focusing on objectives for predictive model information gain. Existing model-based curiosity methods largely rely on ensembles of prediction models to estimate these objectives. In work most similar to our own,
\cite{shyam2018model} computes an estimate of the uncertainty of model predictions derived from the variance of outcomes computed within an ensemble of prediction models. This ensemble-based approach does not scale to compute intensive prediction models. For example, many modern vision-based prediction methods require the full capacity of a GPU to train which would require ensemble based approaches to use many GPUs to explore. This requirement is unreasonable for hardware constrained systems. In this work, we present an objective for scalable exploration based on minimizing a score given by a discriminator network in order to choose actions which result in outcomes considered the least realistic by our adversarial network visualized in Figure \ref{fig:online_method_diagram}.

\textbf{Our Contributions} 
We present an adversarial curiosity reward which we use as an objective in model-based reinforcement learning systems. We integrate this method into two distinct prediction-planning pipelines. In the first pipeline, we perform state-based prediction and plan with a Markov Decision Process to generate exploration and exploitation policies. We demonstrate the scalability of our approach by comparing performance with \cite{shyam2018model} across varying compute restrictions in simulation and show progressively increasing advantages of our method at lower compute targets. 
We also integrate our discriminator with a video prediction model and the cross-entropy method for planning, a compute intensive prediction-planning pipeline for robotic manipulation. This pipeline is trained with randomly sampled data in all prior work and would require hardware modifications on standard robotics platforms to be able to run the approach presented in \cite{shyam2018model}. Our method improved sample efficiency without sacrificing task performance. In addition, our method increased prediction performance. To the best of our knowledge, this application is the first use of model-based curiosity for vision-based robotic manipulation.

Our paper is organized as follows. First, we review the literature on sampling with both curiosity and active learning methods. Then, we present our adversarial curiosity objective for sampling in a model-based reinforcement learning problem. Next, we favorably compare the ability of our method to complete downstream tasks under computation constraints with other curiosity methods using policies learned via exploration. Finally, we demonstrate the ability of our method to scale to a robotic manipulation pipeline in which we show increased sample efficiency and improved prediction performance in a domain transfer problem.

\section{Related Work}
We specify our method as an adversarial curiosity objective which we use to perform active learning in model-based reinforcement learning. Thus, we contextualize our work in the active learning and curiosity literature. The goal of both active learning and curiosity is to select samples with which to train a model such that the model gains the most possible information from the sampled data.

\textbf{Model-Free Curiosity}
Several existing methods use exploration incentives to estimate and seek visual novelty in model-free reinforcement learning. Approaches include rewarding policies based on state visit counts \cite{bellemare2016unifying, ostrovski2017count}, observed error in a prediction model \cite{pathak2017curiosity, burda2018exploration, BucherRSS2019}, disagreement in an ensemble \cite{osband2016deep, Pathak2019}, and information gain in a Bayesian neural network \cite{houthooft2016vime}. We present a model-based curiosity method using the scores from a discriminator network. These same scores could be used as feedback to a model-free method, but we only evaluate the model-based case in this work to highlight the key characteristic of our curiosity approach: scalability.

\textbf{Model-Based Curiosity}
There are far fewer model-based curiosity methods than model-free. Discrete count-based methods \cite{Kearns2002} estimate the learning progress of a potential sample \cite{lopes2012exploration}. \cite{shyam2018model, sekar2020planning} use ensembles of models to estimate uncertainty in predictions. \cite{bechtle2019curious} minimizes the uncertainty of Bayesian models. Our work differs from previous model-based active exploration methods in that it is able to operate computationally efficiently in high-dimensional continuous domains through the use of a discriminator which provides scores for the realism of our model predictions.

\textbf{Adversarial Curiosity} The concept of adversarial curiosity was first proposed by \cite{worldDifferentiable} 
prior to the integration of curiosity with planning algorithms. This work suggested that the formulation of the minimax problem presented a method of encoding introspective behavior in a model. \cite{minimax} further argues that minimax optimization problems such as the one we propose provide intrinsic motivation for a model to invent novel information about which to learn. This behavioral paradigm is considered a form of curiosity \cite{funTheory, worldDifferentiable, minimax, jaegle2019visual}.  We propose the first explicit model of adversarial curiosity with experimental evaluation in this work.

\textbf{Active Learning}
Active learning is the process where a machine learning algorithm selects its training data to improve its data efficiency and performance \cite{settles.tr09}. Many methods for active learning have been proposed, including sampling the most uncertain data points \cite{lewis1994sequential, joshi2009multi, yang2015multi}, sampling where an ensemble of models disagrees \cite{seung1992query, Cohn1994, melville2004diverse}, sampling data that will cause the largest expected information gain \cite{houlsby2011bayesian}, sampling data that will cause the largest expected change in the model \cite{settles2008multiple, sznitman2010active, vezhnevets2012weakly}, sampling data that will cause the largest expected reduction in variance \cite{Cohn1996}, and sampling data that will cause the largest estimated error reduction \cite{roy2001toward, moskovitch2007improving, konyushkova2017learning}. Our approach is most similar to the works that sample the most uncertain data points. To distinguish active learning from curiosity, active learning is specifically formulated to determine samples to select \textit{before} sampling whereas the curiosity literature includes both active methods \cite{bechtle2019curious} and methods in which rewards can only be computed \textit{after} sampling \cite{Pathak2019, pathak2017curiosity}. Thus, the method we propose in this work is both a method for curiosity and active learning.


\section{Adversarial Curiosity Objective}\label{sec:method}

Consider the dynamics of a system $\mathcal{F}$ mapping past states $s_t$ and actions $a_t$ to future states via
\begin{equation}\label{eq:system}
    s_{t+1:t+T_f+1}=\mathcal{F}\left(s_{t-T_p:t},a_{t:t+T_f}\right),
\end{equation}
for $T_f$, $T_p>0$ future and past time intervals, respectively. We then denote by
\begin{equation}
    \textbf{x} = \left(s_{t-T_p:t}, a_{t:t+T_f}, s_{t+1:t+T_f+1}\right)
\end{equation}
any trajectory generated by system \eqref{eq:system}, and denote by $p(\textbf{x})$ the distribution over these trajectories.


Our model-based curiosity method is defined in terms of the following three components:

\begin{enumerate}[i)]
    \item A model $\mathcal{M}$ generates predictions of future states $\widehat{s}$ given past states $s$ and actions $a$. These predictions are made over a prediction horizon $H$ using a set number of past context states $C$. Thus, our prediction model is given by 
    \begin{equation}\label{eq:model}
        \widehat{s}_{t+1:t+H+1} = \mathcal{M}(s_{t-C:t}, a_{t:t+H}).
    \end{equation}
     To lighten notational burden going forward, we let $\textbf{a} := a_{t:t+H}$, $\textbf{c} := s_{t-C:t}$.
    \item A discriminator $\mathcal{D}$, which assigns a score $d_{t}$ to each real trajectory $\textbf{x}$ generated by system \eqref{eq:system} as well as imagined trajectories generated by the prediction model \eqref{eq:model}. To train the discriminator $\mathcal D$, we solve the minimax optimization problem
\begin{align} \label{eq:minimax}
    \min_{\mathcal{M}}&
    \max_{\mathcal{D}}  \mathbb{E}_{\textbf{x} \sim p(\textbf{x})}\left[\log \mathcal{D}\left(\textbf{x}\right)\right] \nonumber \\
    & + \mathbb{E}_{(\textbf{c}, \textbf{a}) \sim p(\textbf{x})}\left[\log \left(1 - \mathcal{D}\left(\textbf{c}, \textbf{a},
    \mathcal{M}(\textbf{c}, \textbf{a}) \right)\right)\right].
\end{align}

The first term in the objective function of optimization problem \eqref{eq:minimax} captures the ability of the discriminator to identify realistic trajectories generated by system \eqref{eq:system}, whereas the second term simultaneously reflects the predictive ability of the model $\mathcal M$, as well as the ability of the discriminator $\mathcal D$ to distinguish between real and imagined trajectories.

The inner maximization trains the discriminator $\mathcal D$ to differentiate between trajectories sampled from the data distribution $\mathbf x$ and predicted trajectories $(\mathbf c, \mathbf a, \mathcal M(\mathbf c, \mathbf a))$. The outer minimization optimizes the performance of the prediction model $\mathcal{M}$. In summary, this minimax problem sets up a competition in which the prediction model tries to learn to make good enough predictions to fool the discriminator while the discriminator tries to improve differentiation of predictions from data samples.

After $\mathcal{D}$ is trained, the discriminator scores for our imagined trajectories are evaluated as
\begin{equation}
    d_{t} = \mathcal{D}\left(\textbf{c}, \textbf{a}, \mathcal{M}(\textbf{c}, \textbf{a})\right).
\end{equation}

\item With these pieces in place, we can now define the curiosity based optimization problem that we solve in order to select action sequences which optimize a curiosity objective defined in terms of the discriminator score.  In particular, we define a planner $\mathcal{P}$ that selects actions which minimize the discriminator score by solving the optimization problem:
\begin{equation}\label{eq:plan}
 \mathcal{P}(\textbf{c}, \textbf{a}, \mathcal{M}, \mathcal{D})  
    := \argmin_{\textbf{a}} \mathcal{D}\left(\textbf{c}, \textbf{a}, \mathcal{M}(\textbf{c}, \textbf{a})\right)
\end{equation}
\end{enumerate}
It then follows that the actions resulting in the least realistic predictions are selected by the planner defined by optimization problem \eqref{eq:plan}, resulting in qualitatively more \textit{curious} behavior.

\begin{figure}
    \centering
    \includegraphics[width=0.9\columnwidth]{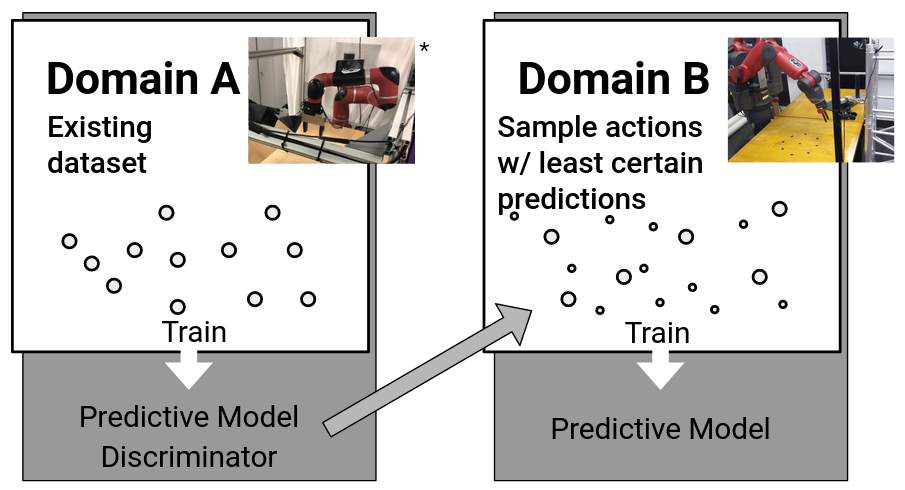}
    \caption{Active sampling to enable domain transfer.  In our robotics experiments, our method trains an action-conditioned prediction model and a discriminator on the dataset in the initial domain.  It then samples actions from the new domain that result in the most uncertain predictions, allowing it to train a prediction model in the new domain with a small number of samples. }
    \label{fig:domain_transfer}

\end{figure}

\begin{figure}
    \centering
    \vspace{0.5cm}
    \includegraphics[width=\columnwidth]{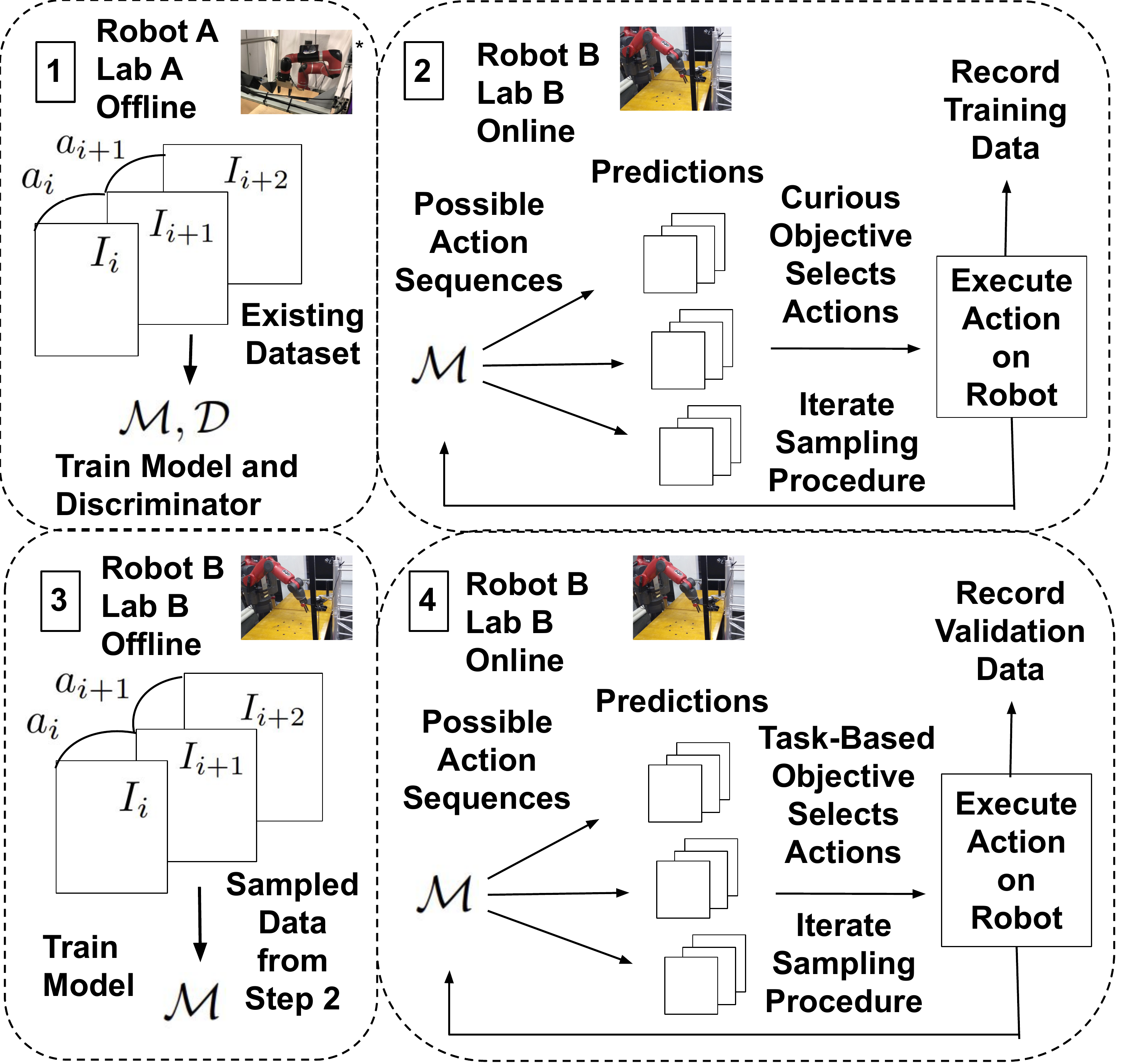}
    \caption{The process used for online training with a curiosity objective provided by the loss from our discriminator network in a domain transfer problem.  The model and the discriminator are initially trained on an existing dataset from domain A (1).  The model and discriminator are used to select and execute sequences of actions that optimize the curiosity objective in domain B, generating a new dataset (2).  The dataset from domain B is used to train the model (3).  The model is used to select sequences of actions that optimize a task-based objective, allowing the robot to perform useful tasks in domain B (4).}
    \label{fig:exp_setup}
    \vspace{-0.25cm}
\end{figure}

We note that in the domain transfer problem visualized in Figure~\ref{fig:domain_transfer} introduces a variant on this process for sampling. The model $\mathcal{M}$ is first trained jointly with the discriminator $\mathcal{D}$ on data from Domain A. Then, the model $\mathcal{M}$ and the discriminator $\mathcal{D}$ are used in the planner $\mathcal{P}$ to execute the sampling procedure in Domain B in order to gather data for updating $\mathcal{M}$. If the discriminator will continue to be used for future collection tasks, $\mathcal{D}$ can be trained jointly with $\mathcal{M}$ again to be updated using the newly sampled data. This sampling procedure for domain transfer is laid out in more detail for our specific experimental application in Figure~\ref{fig:exp_setup}.

\section{Exploration Results Under Computation Constraints}

\begin{figure*}
    \centering
    \vspace{0.1cm}
    \includegraphics[width=\textwidth]{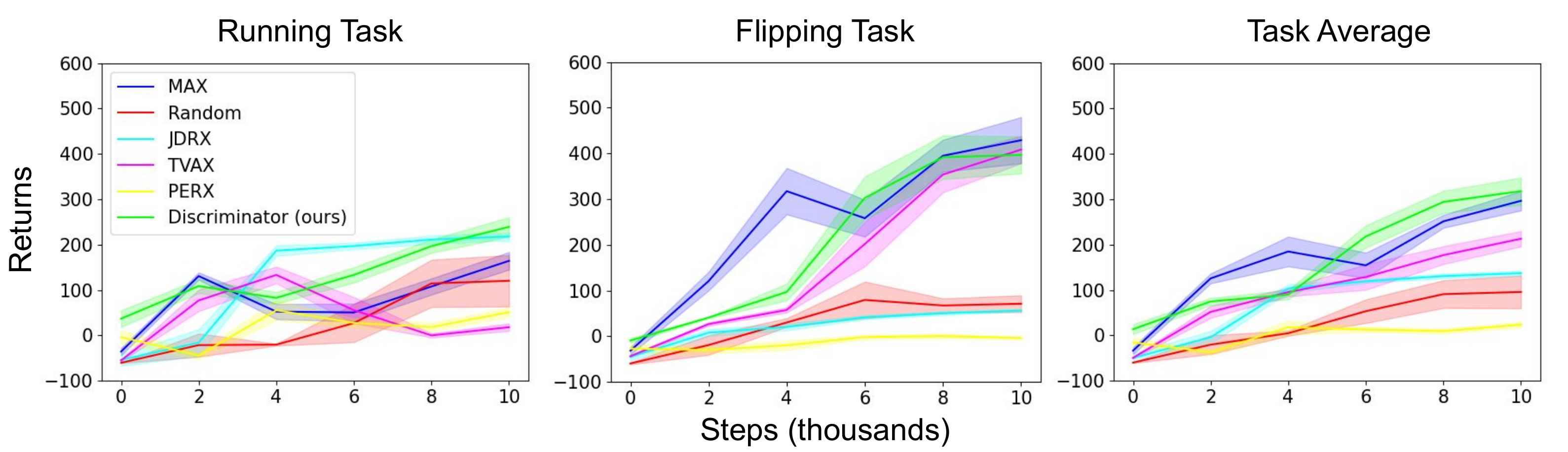}
    \caption{Extrinsic rewards from running and flipping task completion in the Half Cheetah OpenAI environment averaged over 5 trials initiated with random seeds broken down by each distinct method for exploration used in training. All ensemble based curiosity methods (MAX, JDRX, TVAX) were trained with an ensemble size of 32.
    Our approach outperforms competing curiosity methods by exploring in a way that allows it to perform well on both tasks.
    }
    \label{fig:baselines}
\end{figure*}

\begin{figure*}
    \centering
    \includegraphics[width=\textwidth]{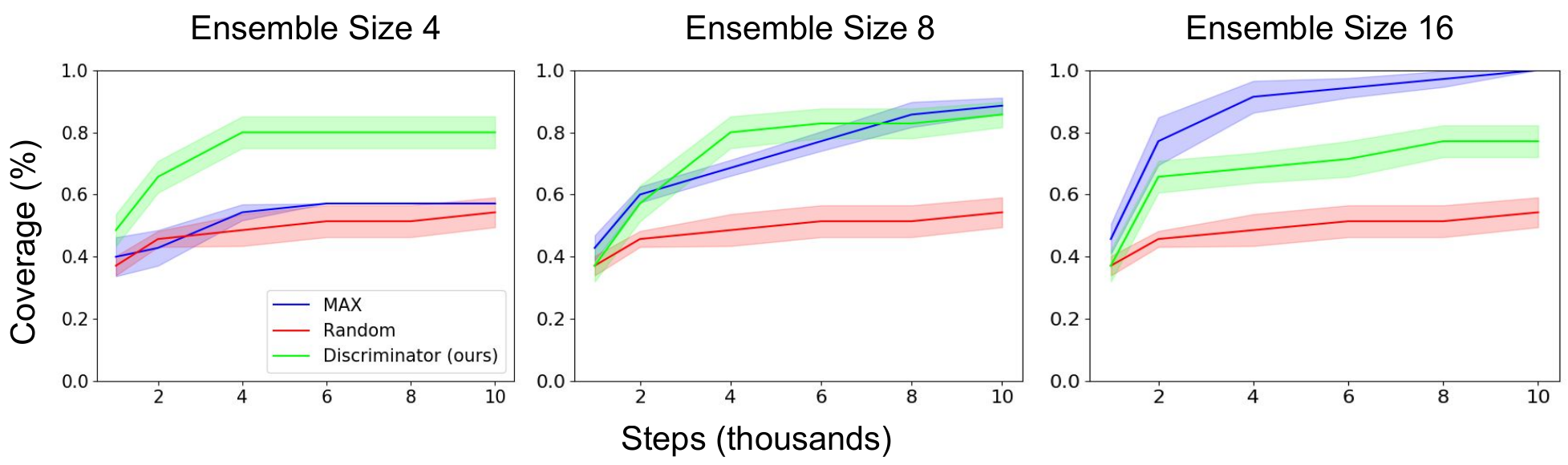}
    \caption{Maze coverage results in the Ant Maze OpenAI environment averaged over 5 trials initiated with random seeds. Our approach outperforms ensemble-based methods when low ensemble sizes are used due to computation constraints. Perception models in robot learning applications often use the full capacity of a GPU to train, so the size of the ensemble would correspond with the number of needed GPUs making ensemble-based exploration strategies computationally intractable.}
    \label{fig:scalability}
\end{figure*}

We compare our curiosity method to two model-based exploration methods both presented in \cite{shyam2018model}, the closest work to our own. Model-based active exploration (MAX) rewards exploration with the Jensen-Shannon or Jensen-R\'{e}nyi divergence between an ensemble of prediction models to measure prediction uncertainty. The Jensen-R\'{e}nyi divergence is used for continuous state spaces as is the case in our experiments. Trajectory variance active exploration (TVAX) computes the variance in sampled trajectories between an ensemble of prediction models as an exploration reward. We also compare our method to the model-free prediction error approach presented in \cite{pathak2017curiosity} (PERX) and a model-free version of MAX (JDRX) as well as to random exploration.

In order to execute a fair comparison, we use the same prediction-planning pipeline proposed with the MAX algorithm across all of our simulated experiments so that performance differences can be based solely on the different exploration policy approaches. All methods use an ensemble of fully connected neural networks trained with a negative log-likelihood loss jointly with our discriminator to predict next state distributions. Next state predictions are given by the distribution mean. We emphasize that this use of ensembles as a predictive model is not necessary in the case of our curiosity method other than to standardize the prediction approach for comparison purposes whereas MAX, JDRX, and TVAX all require an ensemble-based prediction approach to derive their curiosity rewards. Similarly, the discriminator is only used for our curiosity method but trained with every method to standardize the prediction approach. The model-free methods use soft-actor critic (SAC)~\cite{haarnoja2018soft} to learn a policy.
The model-based methods are trained with data gathered by an MDP using an exploration objective following the framework presented in \cite{shyam2018model}.

We execute our first set of experiments in the OpenAI Half Cheetah environment which has an 18-dimensional continuous observation space and a 6-dimensional continuous action space~\cite{brockman2016openai}.
Following \cite{shyam2018model}, we add Gaussian noise of $\mathcal{N}(0, 0.02)$ to the actions to increase the difficulty.
Policies are trained purely with each exploration strategy and then used to execute downstream tasks.
We compare the performance of each exploration method by evaluating achieved extrinsic rewards in these tasks over 5 randomly seeded trials and present them in Figure~\ref{fig:baselines}. We find that our method either slightly outperforms or performs comparably to all the baselines in both downstream tasks: running and flipping. More significantly, our approach slightly outperforms competing curiosity methods by exploring in a way that allows it to perform well on both tasks, achieving the highest average task performance.

Next, we compare maze coverage results in the OpenAI Ant Maze environment in our experiments presented in Figure~\ref{fig:scalability}. We note that the key contribution of our approach comes from the scalability of the discriminator. We observe that at high ensemble sizes (16 or more prediction models), MAX outperforms our discriminator. However, as ensemble size decreases, our method significantly outperforms MAX. Thus, we show that when compute restrictions prevent the use of large ensembles of prediction models, the performance of ensemble based methods are significantly compromised while our method continues to perform well. We demonstrate the significance of this result in Section~\ref{sec:testing_procedure} where we integrate our method in an established prediction-planning pipeline for robotic manipulation for which hardware changes would be required to run an ensemble based approach.

\footnotetext[1]{Image used with permission from \cite{Xie2019}}

\section{Robotic Manipulation}\label{sec:testing_procedure}

We demonstrate the ability of our adversarial curiosity method to meaningfully scale to robotics problems by integrating our approach in a prediction-planning pipeline commonly studied robot learning approaches to manipulation. Robot learning methods already face a scalability problem, as large amounts of data are required for training which is time consuming to collect. To address this challenge, recent research demonstrated the ability of a class of these data-driven methods to transfer across platforms \cite{DasariCoRL2019}. This capability allows for publicly available datasets to be leveraged for the bulk of the training, and only a small amount of data needs to be collected for fine-tuning in the new domain. We propose performing targeted sampling with our adversarial curiosity objective in the new domain instead of the random sampling used in prior work, as illustrated in Figure~\ref{fig:domain_transfer}. We demonstrate increased sample efficiency and gains in prediction performance from incorporating our discriminator in this pipeline.

States (denoted $s_t$ in prior sections) are RGB images in this set of experiments, so we specify our notation by referencing these image states as $I_t$. We use a variant of the prediction model from \cite{DasariCoRL2019}. A stack of convolutional LSTMs is used to predict a flow field from an image $I_t$ and action $a_t$. This flow field is then applied directly to the input image $I_t$ to predict the next image frame $\widehat{I}_{t+1}$.
The true next image frame $I_{t+1}$ is observed after the given action is taken. This network is optimized with an $L_1$ loss between the predicted image $\widehat{I}_{t+1}$ and true image $I_{t+1}$. In practice, these models perform predictions out to some horizon $H$ using a context of $C$ image frames in which the flow field estimates are applied recursively across the prediction horizon.

We extend the notation presented in Section \ref{sec:method} by setting 
\begin{equation}
    \textbf{h} = I_{t+1:t+H+1}
\end{equation}
such that a sampled trajectory is given by
\begin{equation}
\textbf{x}=\left(\textbf{c}, \textbf{a}, \textbf{h}\right) = \left(I_{t-C:t}, a_{t+1:t+H+1}, I_{t+1:t+H+1}\right).
\end{equation}

In the training procedure described by Step 1 in Figure \ref{fig:exp_setup}, our prediction model is optimized jointly with the discriminator defined in Section \ref{sec:method}. The optimization problem solved by our model $\mathcal{M}$ during training is
\begin{align}\label{eq:total_loss}
     \min_{\mathcal{M}}\max_{\mathcal{D}} & \mathbb{E}_{\textbf{x} \sim p(\textbf{x})}\left[L_1\left(\textbf{h}, \mathcal{M}\left(\textbf{c}, \textbf{a}\right)\right)\right] + \mathbb{E}_{\textbf{x} \sim p(\textbf{x})}\left[\log \mathcal{D}\left(\textbf{x}\right)\right] \nonumber \\ 
    & + \mathbb{E}_{(\textbf{c}, \textbf{a}) \sim p(\textbf{x})}\left[\log \left(1 - \mathcal{D}\left(\textbf{c}, \textbf{a},
    \mathcal{M}(\textbf{c}, \textbf{a}) \right)\right)\right].
\end{align}
which combines the adversarial minimax game from equation \eqref{eq:minimax} with the L1 loss on prediction error.  This is similar to the loss in \cite{Lee2018}, where the combination of prediction error and an adversarial loss were shown to improve prediction quality and convergence.

We use this prediction model and the cross-entropy method (CEM) \cite{Rubinstein1999} to optimize for a sequence of actions that minimize Equation~\ref{eq:plan}.
This procedure causes the robot to select a sequence of actions that generate unrealistic looking predictions.

\begin{figure}
    \centering
    \includegraphics[width=\columnwidth]{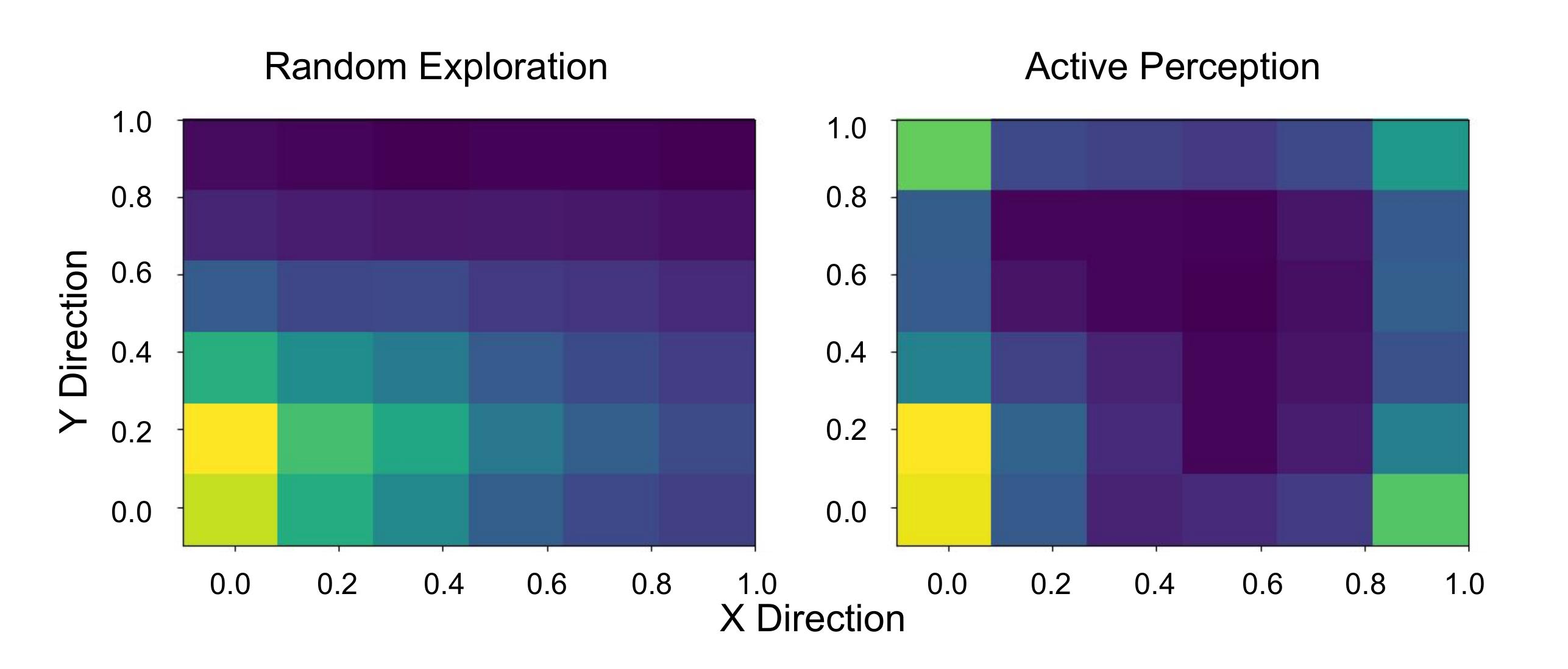}
    \caption{Heat map of the state space regions explored by each policy over 650 trajectories. Regions which are more yellow indicate a higher count for the end effector of the Baxter arm accessing that discretized x-y region.  Our curious model explores all corners of the state space, focusing on the edges where objects accumulate, while the random exploration remains near its starting location.}
    \label{fig:statespace}
\end{figure}


\subsection{Sampling Analysis}\label{sec:sampling}

We evaluate the ability of our curiosity objective to effectively explore the environment by comparing the behavior of our curious policy to the behavior of the random policy used in prior work. To make this comparison, we execute Steps 1 and 2 visualized in Figure \ref{fig:exp_setup}. First, our prediction model and discriminator is jointly trained on Sawyer data from the RoboNet dataset \cite{DasariCoRL2019} by optimizing equation \eqref{eq:total_loss}. Then, we use each policy to separately sample trajectories on a Baxter robot platform. Our curious policy was able to visit a more diverse array of states and grasp more objects than the existing random policy.

\begin{figure}
    \centering
    \includegraphics[width=0.9\columnwidth]{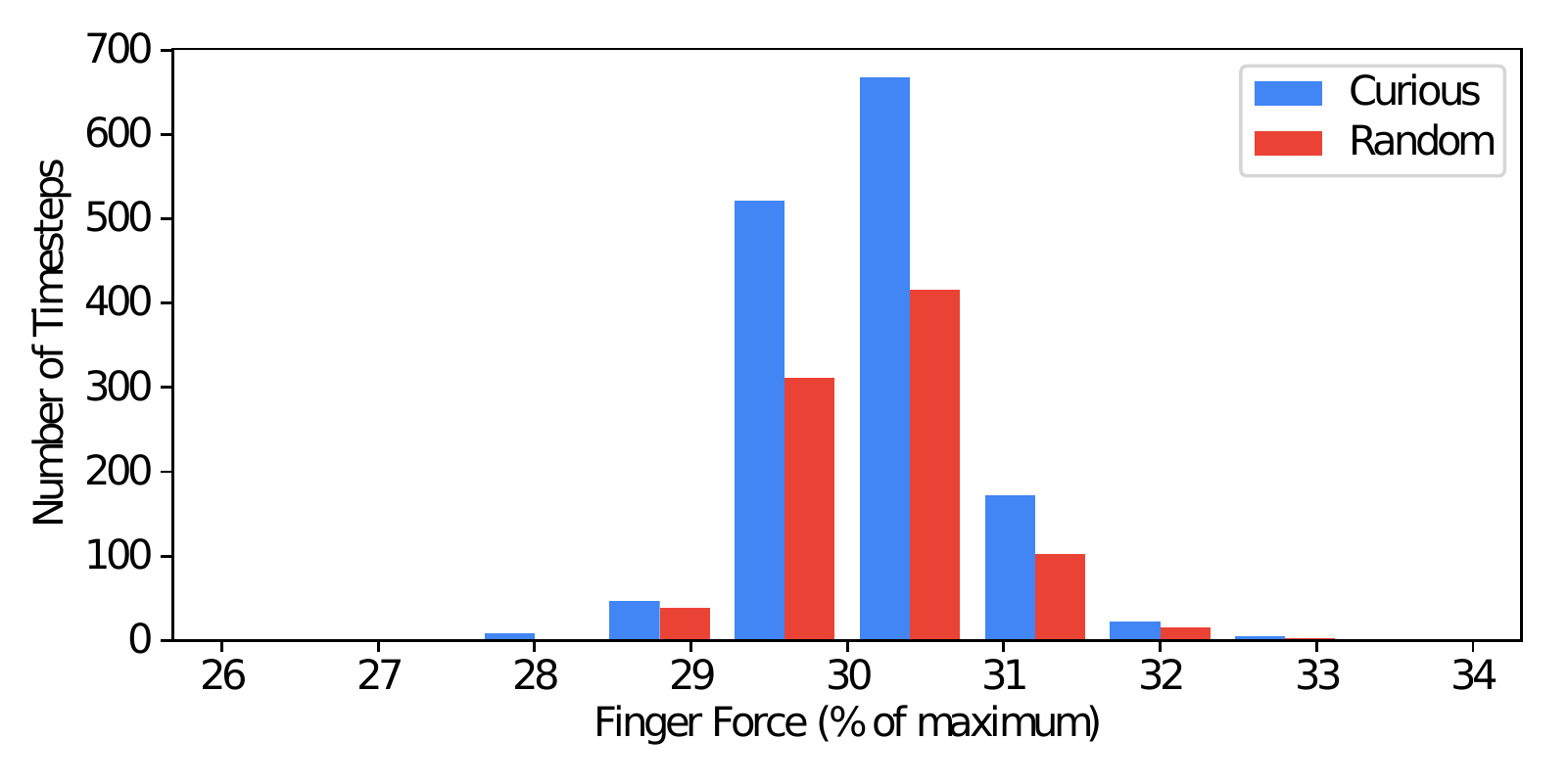}
    \caption{Histograms of the non-zero gripper forces experienced while executing each policy.  Each policy was executed for 650 trajectories of 30 timesteps.  Non-zero force occurs when a large enough object is grasped by the robot's fingers.  The curious policy spends significantly more time grasping objects than the random policy.}
    \label{fig:grippers}
\end{figure}

Our policy is able to significantly increase the quantity of objects that the robot grasps.  Since the interaction between the robot and objects are some of the most difficult things to predict in the tabletop manipulation setting, having more data about robot-object interactions makes a collected dataset more effective in training prediction models. 
Figure~\ref{fig:grippers} shows a histogram of when the grippers of the robot experienced non-zero forces during data collection for both the curious and the random policies.  Non-zero forces indicate that an object is between the grippers, preventing them from fully closing.  When following the curious policy, the robot spends a larger portion of its time grasping objects.

Our curious policy also explores a 
different distribution of robot states.
Figure~\ref{fig:statespace} shows a heatmap of the amount of time the robot's end effector spends at each location in the xy-plane.  The curious policy explores more interesting regions of the state space, such as edges of the bins.  The walls of the bin are interesting because they block the motion of objects, causing the objects to have more complicated dynamics than when they are in the center of the bin.


\subsection{Prediction Performance Improvement}

We also evaluated the ability of the model trained with data collected under different policies to perform prediction for robotic control by executing Steps 3 and 4 visualized in Figure \ref{fig:exp_setup}. 
\begin{figure}
\input{tex/diagrams/prediction_results.tex}
\begin{subfigure}{0.45\textwidth}
    \centering
    \includegraphics[width=\textwidth]{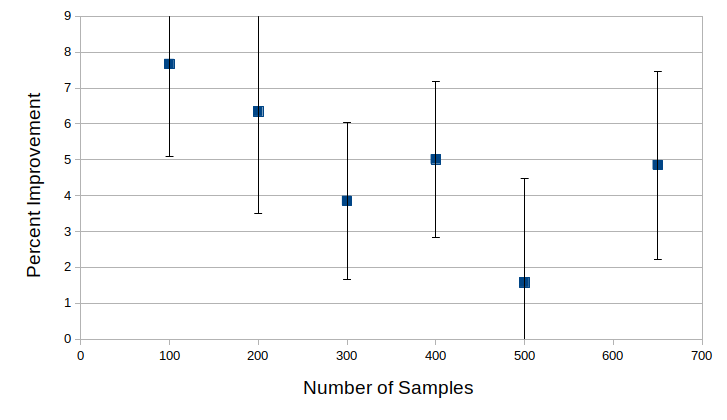}
    \caption{Percent improvement in L2 error for the prediction model trained with curious data over the prediction model trained with the random data on the control dataset.  The prediction model trained with curious data performs better by more than the standard error on all but one quantity of samples. The improved performance is particularly pronounced when lower numbers of samples are used in training.}
    \label{fig:improvement}
\end{subfigure}
\caption{Qualitative and quantitative prediction results.}
\end{figure}
The samples collected with our curious policy enable statistically significant prediction improvement on control tasks over samples collected with our random policy.
The $L_2$ error improvement for the model trained with the data collected with the curious policy at different numbers of samples is visualized in Figure~\ref{fig:improvement}.
Error improvement for the curious policy is especially pronounced at lower numbers of samples.
Qualitative prediction results are shown in Figure~\ref{fig:robotic_predictions}.  The model trained with the curious data more accurately tracks the position of the object.

Given an improved prediction model, we experimented with planning performance.
We use CEM for planning with a task-based objective function first proposed by \cite{Finn2016}, using the exact formulation and implementation from \cite{DasariCoRL2019}. 
Despite the significant improvement in our prediction model, We found that the models trained with curious and random data resulted comparable control performance. In future work, we intend to investigate the relationship between prediction and control performance in our experiments.





\section{Conclusion}
We presented an adversarial curiosity approach which we use to actively sample data used to train a prediction model. Our method optimizes an objective given by the score from a discriminator network to choose the sequence of actions that corresponds to the least realistic sequence of predicted observations. 
We favorably compared our approach in simulated task execution to existing model-based curiosity methods and demonstrated significant advantages of our method in the case of compute constrained environments.
We then integrated our exploration strategy in a practical robotic manipulation pipeline where we demonstrated increased sample efficiency and improved prediction performance in a domain transfer problem. 
In future work, we plan to look into ways to guarantee improvement with active exploration methods for manipulation task performance in robotic learning pipelines.

\section*{Acknowledgements}
The authors are grateful for support through the Curious Minded Machines project funded by the Honda Research Institute.

\clearpage

\bibliographystyle{IEEEtran}
\bibliography{IEEEabrv,example}

\end{document}